%% file: main.tex
\pdfoutput=1

\documentclass[11pt]{article}

\usepackage{natbib}
\usepackage{acl}
\usepackage{xcolor}	

\usepackage{xurl}
\usepackage{times}
\usepackage{latexsym}

\usepackage[T1]{fontenc}

\usepackage[utf8]{inputenc}

\usepackage{microtype}

\usepackage{graphicx}
\usepackage{amsmath,amsthm,amssymb}
\usepackage{mathtools}
\usepackage{booktabs}
\usepackage{enumitem}
\usepackage{bm}
\usepackage{comment}
\usepackage{dblfloatfix}
\usepackage{subcaption}
\usepackage{pbox}

\usepackage{colortbl} 

\newcolumntype{R}[1]{>{\raggedleft\arraybackslash}p{#1}} 

\title{Do Transformer Models Show Similar Attention Patterns to\\ Task-Specific Human Gaze?}

\author{Stephanie Brandl$^{*,1,2}$, Oliver Eberle$^{*,1}$, Jonas Pilot$^{1}$,  Anders Søgaard$^{2}$ \\
  $^{1}$Machine Learning Group, TU Berlin, Germany \quad
  $^{2}$University of Copenhagen, Denmark\\
\texttt{oliver.eberle@tu-berlin.de}, \texttt{\{brandl, soegaard\}@di.ku.dk} \\
$^{*}$Authors contributed equally.
  }
  
\begin{document}
\maketitle
\begin{abstract}
Learned self-attention functions in state-of-the-art NLP models often correlate with human attention. We investigate whether self-attention in large-scale pre-trained language models is as predictive of human eye fixation patterns during task-reading as classical cognitive models of human attention. 
We compare attention functions across two task-specific reading datasets for sentiment analysis and relation extraction. 
We find the predictiveness of large-scale pre-trained self-attention for human attention depends on `what is in the tail', e.g., the syntactic nature of rare contexts.
Further, we observe that task-specific fine-tuning does not increase the correlation with human task-specific reading. Through an input reduction experiment we  give complementary insights on the sparsity and fidelity trade-off, showing that lower-entropy attention vectors are more faithful.

\end{abstract}

\section{Introduction}
The usefulness of learned self-attention functions often correlates with how well it aligns with human attention \cite{das-etal-2016-human,klerke-etal-2016-improving,barrett-etal-2018-sequence,zhang-zhang-2019-using,klerke-plank-2019-glance}. In this paper, we evaluate how well attention flow \cite{abnar-zuidema-2020-quantifying} in large language models, namely BERT \cite{devlin-etal-2019-bert}, RoBERTa \cite{liu2019roberta} and T5 \cite{JMLR:v21:20-074}, aligns with human eye fixations during task-specific reading, compared to other shallow sequence labeling models \cite{lecun95convolutional,vaswani2017attention} and a classic, heuristic model of human reading \cite{reichle03ez}. We compare the learned attention functions and the heuristic model across two task-specific English reading tasks, namely sentiment analysis (SST movie reviews) and relation extraction (Wikipedia), as well as natural reading, using a publicly available dataset with eye-tracking recordings of native speakers of English \cite{hollenstein2018zuco}. 

\paragraph{Contributions}
We compare human and model attention patterns on both sentiment reading and relation extraction tasks. In our analysis, we compare human attention to pre-trained Transformers (BERT, RoBERTa and T5),  from-scratch training of two shallow sequence labeling architectures \cite{lecun95convolutional,vaswani2017attention}, as well as to a frequency baseline and a heuristic, cognitively inspired model of human reading called the {E-Z~Reader} \cite{reichle03ez}. We find that the heuristic model correlates well with human reading, as has been reported in \citet{NEURIPS2020_460191c7}. However when we apply {\em attention flow} \cite{abnar-zuidema-2020-quantifying}, the pre-trained Transformer models also reach comparable levels of correlation strength.
Further fine-tuning experiments on BERT did not result in increased correlation to human fixations. To understand what drives the differences between models, we perform an in-depth analysis of the effect of word predictability and POS tags on correlation strength. It reveals that Transformer models  do not accurately capture tail phenomena for hard-to-predict words (in contrast to the E-Z Reader) and that Transformer attention flow shows comparably weak correlation on (proper) nouns while the E-Z Reader predicts importance of these more accurately, especially on the sentiment reading task. In addition, we investigate a subset of the ZuCo corpus for which aligned task-specific and natural reading data is available and find that Transformers  correlate stronger to natural reading patterns.
We test faithfulness of these different attention patterns to
produce the correct classification via an input reduction experiment on task-tuned BERT models. Our results highlight the trade-off between model faithfulness and sparsity when comparing importance scores to human attention, i.e., less sparse (higher entropy) attention vectors tend to be less faithful with respect to model predictions. Our code is available at \href{https://github.com/oeberle/task_gaze_transformers}{\nolinkurl{github.com/oeberle/task_gaze_transformers}}.

\section{Pre-trained Language Models vs Cognitive Models}

\citet{church21future} discuss how NLP has historically benefited from rationalist and empiricist methodologies, something that holds for cognitive modeling in general. The vast majority of application-oriented work in NLP today relies on pre-trained language models or other large-scale data-driven models, but in cognitive modeling, most approaches remain heuristic and rule-based, or hybrid, e.g., relying on probabilistic language models to quantify surprisal \cite{rayner10models,milledge19changing}. This is for good reasons: Cognitive modeling values interpretability (even) more, often suffers from data scarcity, and is less concerned with model reusability across different contexts. 

This paper presents a head-to-head comparison of the {E-Z~Reader} and pre-trained Transformer-based language models. We are not the first to evaluate pre-trained language models and large-scale data-driven models as if they were cognitive models. \citet{chrupala-alishahi-2019-correlating}, for example, use representational similarity analysis to correlate sentence encodings in pre-trained language models with fMRI signals; \citet{abdou-etal-2019-higher} correlate sentence encodings with gaze-derived representations. More generally, it has been argued that cognitive evaluations are in some cases practically superior to standard evaluation methodologies in NLP \cite{sogaard-2016-evaluating,hollenstein-etal-2019-cognival}. We return to this in the Discussion and Conclusion \S\ref{sec:disc}. 

Commonly, pre-trained language models are disregarded as cognitive models, since they are most often implemented as computationally demanding batch learning algorithms, processing data “at once”. \citet{gunther19vector} points out that this is an artefact of their implementation, and online learning of pre-trained language models is possible, yet impractical. Generally, several researchers have argued for taking pre-trained language models seriously as cognitive models \cite{rogers2016conceptual,mandera17explaining,gunther19vector}. 
In the last section, \S\ref{sec:disc}, 
we discuss some of the implications of comparisons of pre-trained language models and cognitive models -- for cognitive modeling, as well as for NLP. 
In our experiments, we focus on Transformer architectures that are currently the dominating pre-trained language models and a {\em de facto}~baseline for modern NLP research. 

\vspace{-2mm}
\begin{figure*}[th!]
\includegraphics[width=1.0\linewidth]{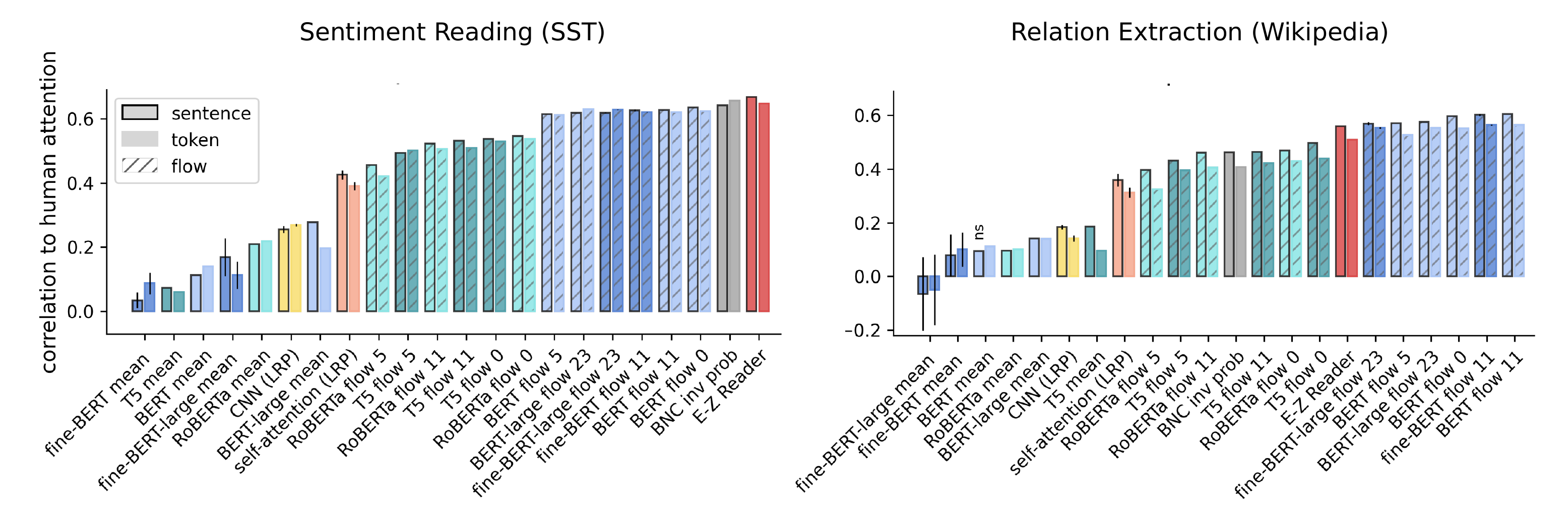}
\caption{Spearman correlation analysis between human attention and different models for two task settings. Solid bar edges indicate sentence-level correlations in contrast to a token-level analysis. \textit{Left:} Sentiment Reading on the SST dataset. \textit{Right:} Relation Extraction on Wikipedia. Standard deviations over five seeds are shown for fine-tuned models and correlations are statistically significant with $p < 0.01$ unless stated otherwise (ns: not significant).}
\label{fig:spearman}
\end{figure*}

\section{Experiments}
\subsection{Data}
The ZuCo dataset \cite{hollenstein2018zuco} contains eye-tracking data for 12 participants (all English native speakers) performing natural reading and relation extraction on 300 and 407 English sentences from the Wikipedia relation extraction corpus \cite{culotta2006integrating} respectively and sentiment reading on 400 samples of the Stanford Sentiment Treebank (SST) \cite{socher2013recursive}. For our analysis, we extract and average word-based total fixation times across participants and focus on the task-specific relation extraction and sentiment reading samples. 

\subsection{Models}
Below we briefly describe our used models and refer to Appendix \ref{app:model_optimization} for more details.

\paragraph{Transformers} The superior performance of Transformer architectures across broad sets of NLP tasks raises the question of how task-related attention patterns really are.  In our experiments, we focus on comparing task-modulated human fixations to  attention patterns extracted from the following commonly used models: (a) We use both pre-trained uncased BERT-base and large models \cite{devlin-etal-2019-bert} as well as fine-tuned BERT models on the respective tasks. BERT was originally pre-trained on the English Wikipedia and the BookCorpus. (b) The RoBERTa model has the same architecture as BERT and demonstrates better performance on downstream tasks using an improved pre-training scheme and the use of additional news article data \cite{liu2019roberta}. (c) The Text-to-Text Transfer Transformer (T5) uses an encoder-decoder structure to enable parallel task- training and has demonstrated state-of-the-art performance over several transfer tasks including sentiment analysis and natural language inference \cite{JMLR:v21:20-074}.

We evaluate different ways of extracting token-level importance scores: 
We collect attention representations and compute the mean attention vector over the final layer heads to capture the mixing of information in Transformer self-attention modules as in \citet{hollenstein-beinborn-2021-relative} and present this as \textit{mean} for all aforementioned Transformers.

To capture the layer-wise structure of deep Transformer models we compute attention flow \cite{abnar-zuidema-2020-quantifying}.
This approach considers the attention matrices as a graph, where tokens are represented as nodes and attention scores as edges between consecutive layers. The edge values define the maximal flow possible between a pair of nodes. Flow between edges is thus (i) limited to the maximal attention between any two consecutive layers for this token  and (ii) conserved such that the sum of incoming flow must be equal to the sum of outgoing flow. We denote the attention flow propagated back from  layer $L$ as {\em flow $L$}.

\paragraph{Shallow Models}
We ground our analysis on Transformers by comparing them to relatively shallow models that were trained from-scratch and evaluate how well they coincide with human fixation.
We train a standard \textbf{CNN} \cite{kim-2014-convolutional} network with multiple filter sizes on pre-trained GloVe embeddings \cite{pennington2014glove}. Importance scores over tokens are extracted using Layerwise Relevance Propagation (LRP) \cite{arras-etal-2016-explaining, arras2017_what_is_relevant} which has been demonstrated to produce robust explanations by iterating over layers and redistributing relevance from outer layers towards the input \cite{bach-plos15, XAIreview2021}.
In parallel, we use a shallow multi-head \textbf{self-attention} network 
\cite{DBLP:journals/corr/LinFSYXZB17} on GloVe vectors with a linear read-out layer for which we compute token relevance scores using LRP.

\paragraph{E-Z~Reader}
As a cognitive model for human reading, we compute task-neutral fixation times using the E-Z~Reader \cite{reichle1998toward} model. The E-Z~Reader is a multi-stage, hybrid model, which relies on an $n$-gram model and several heuristics, based, for example, on theoretical assumptions about the role of predictability and average saccade length. Additionally, we compare to a frequency baseline using word statistics of the \textbf{BNC} (British National Corpus, \cite{BNCKilgarriff})\footnote{We compute the negative log-transformed probability of each lower-cased token corresponding to an inverse relation between word-frequency and human gaze duration \cite{PMID:3736392} as proposed by \citet{barrett-etal-2018-sequence}.} 

\subsection{Optimization}
For training models on the different tasks we remove all sentences that overlap between ZuCo and the original SST and Wikipedia datasets. Models are then trained on the remaining train-split data until early stopping is reached and we report results over five runs. We provide further details on the optimization and model task performance in Appendix \ref{app:model_optimization}.

\subsection{Metric} 
To compare models with human attention, we compute Spearman correlation between human and model-based importance vectors after concatenation of individual sentences as well as on a token-level, see \citet{hollenstein-beinborn-2021-relative}. This enables us to distinguish unrelated effects caused by varying sentence length from token-level importance. 
As described before, we extract human attention from gaze (ZuCo), simulated gaze (E-Z Reader), raw attentions (BERT, RoBERTa, T5), relevance scores (CNN, self-attention) and inverse token probability scores (BNC).\footnote{First and last token bins from each sentence are ignored to avoid the influence of sentence border effects in Transformers \cite{clark-etal-2019-bert} and for which the E-Z Reader does not compute fixations.} We use ZuCo tokens to align sentences across tokenizers and apply max-pooling of scores when bins are merged. 

\begin{figure*}[t!]
    \centering
    \includegraphics[width=\textwidth]{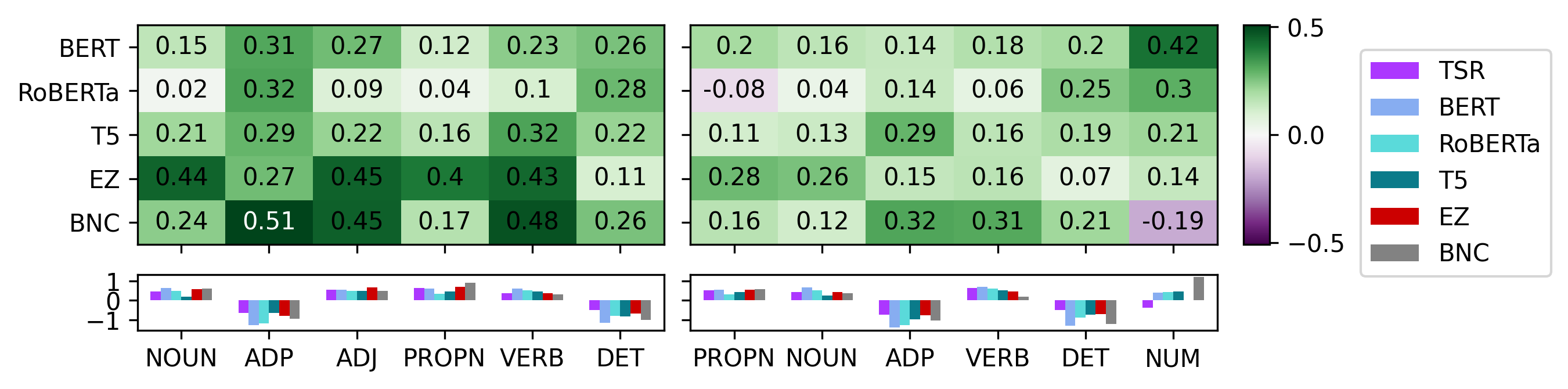}
    \caption{\textit{Upper:} Correlations between human fixation and different models for SST \textit{(left)} and Relation Extraction \textit{(right)} for the six most common POS tags. \textit{Lower:} Average attention value after standardization (mean=0, std=1) for respective POS tag and model.}
    \label{fig:pos}
\end{figure*}

\subsection{Main result}
To evaluate how well model and human attention patterns for sentiment reading and relation extraction align, we compute pair-wise correlation scores as displayed in Figure \ref{fig:spearman}.
Reported correlations are statistically significant with $p < 0.01$ if not indicated otherwise (ns: not significant). After ranking based on the correlations on sentence-level, we observe clear differences between sentiment reading on SST and relation extraction on Wikipedia for the different models. For sentiment reading, the {E-Z~Reader} and BNC show the highest correlations followed by the Transformer attention flow values (the ranking between E-Z/BNC and Transformer flows is significant at $p<0.05$ ). For relation extraction, we see the highest correlation for BERT-base attention flows (with and without fine-tuning) and BERT-large followed by the {E-Z~Reader} (ranking is significant at $p<0.05$). On the lower end, computing means over BERT attentions across the last layer shows weak to no correlations for both tasks.\footnote{We have experimented with oracle analyses selecting the maximally correlating attention head in the last layer for each sentence and find that correlations are generally weaker than with attention flow. \label{oracle}} The shallow architectures result in low to moderate correlations with a distinctive gap to attention flow. Focusing on flow values for Transformers, BNC and E-Z~Reader, correlations are stable across word and sentence length.  Correlations grouped by sentence length shows stable values around $0.6$ (SST) and $0.4-0.6$ (Wikipedia) except for shorter sentences where correlations fluctuate. To check the linear relationship between human and model attention patterns we additionally compute token- and sentence-level Pearson correlations which can be found in Appendix \ref{app:spearman_pearson}. Results confirm that Spearman and Pearson correlation coefficients as well as rankings hardly differ - which suggests a linear relationship - and that correlation strength is in line with \citet{hollenstein-beinborn-2021-relative}.

\begin{figure*}[t]
    \centering
    \includegraphics[width=\textwidth]{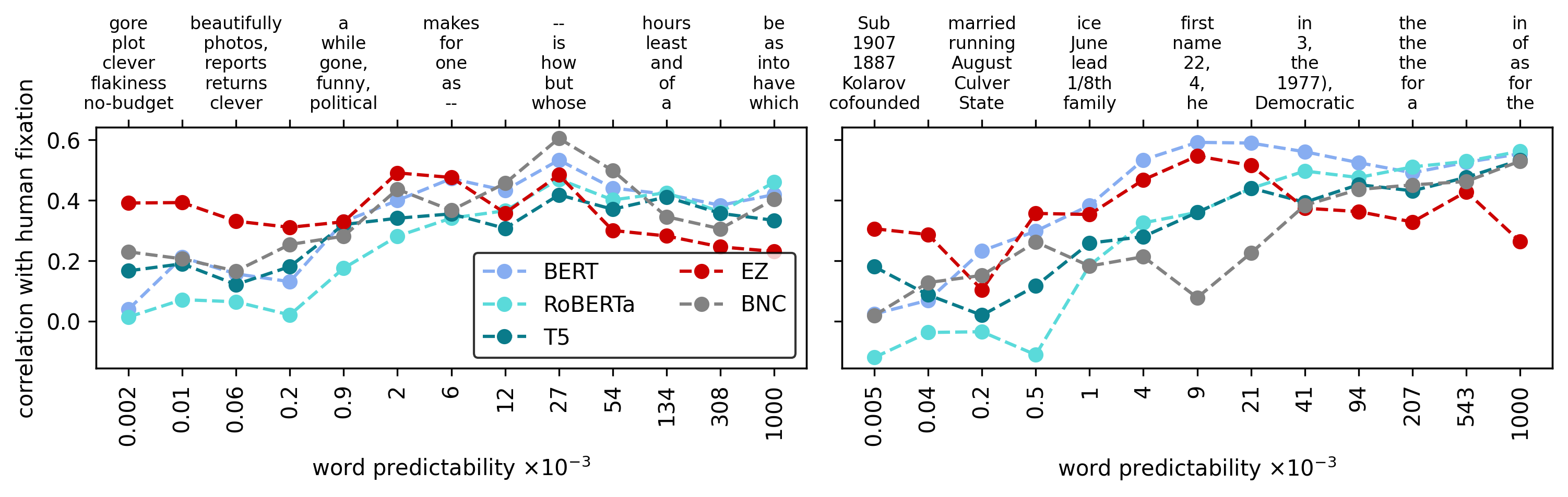}
    \caption{Correlation between human fixations and different models for SST \textit{(left)} and Wikipedia \textit{(right)} with respect to word predictability in equally sized bins. Word predictability scores, were calculated with a 5-gram Kneser-Ney language model. 
    Respective bin limits are given on the x-axis. Samples for every other bin are displayed on the upper x-axis.}
    
    \label{fig:cloze}
\end{figure*}

\section{Analyses}\label{sec:analyses}
In addition to our main result -- that pre-trained language models {\em are}~competitive to heuristic cognitive models in predicting human eye fixations during reading -- we present a detailed analysis, investigating what our main results depend on, where pre-trained language models improve on cognitive models, and where they are still challenged. 

\paragraph{Fine-tuning BERT does not change correlations to human attention} 
We find that fine-tuning base and large BERT models on either task does not significantly change correlations and are of similar strength to untuned models. This observation can be embedded into findings that Transformers are equipped with overcomplete sets of attention functions that hardly change until the later layers, if at all, during fine-tuning and that this change is also dependent on the tuning task itself \cite{kovaleva2019revealing, zhao-bethard-2020-berts}. 
In addition, we observe that Transformer flows propagated back from early, medium and final layers do not considerably change correlations to human attention. This can be explained by attention flow filtering the path of minimal value at each layer as discussed in \citet{abnar-zuidema-2020-quantifying}.

\paragraph{Attention flow is important} 
The correlation analysis emphasizes that we need to capture the layered propagation structure in Transformer models, e.g., by using attention flow, in order to extract importance scores that are competitive with cognitive models. Interestingly, selecting the highest correlating head for the last attention layer produces generally weaker correlation than attention flows.
This offers additional evidence that raw attention weights do not reliably correspond to token relevance \cite{serrano-smith-2019-attention, abnar-zuidema-2020-quantifying} and, thus, are of limited use to compare task attention to human gaze.

\paragraph{Differences between language models} BERT, RoBERTa and T5 are large-scale pretrained language models based on Transformers, but they also differ in various ways. One key difference is that BERT and RoBERTa use absolute position encodings, while T5 uses relative encodings. BERT and RoBERTa differ in that (i) BERT has a next-sentence-prediction auxiliary objective; (ii) RoBERTa and T5 were trained on more data; (iii) RoBERTa uses dynamic masking and trains with larger mini-batches and learning rates, while T5 uses multi-word masking; (iv) RoBERTa uses byte pair encoding for subword segmentation. We leave it as an open question whether the superior attention flows of BERT, compared to RoBERTa and T5, has to do with training data, next sentence prediction, or fortunate hyper-parameter settings, but note that BERT is also known to have higher alignment with human-generated explanations than other large-scale pre-trained language models \cite{prasad-etal-2021-extent}.

\paragraph{E-Z~Reader is less sensitive to hard-to-predict words and POS}
We compare correlations to human fixations with attention flow values for Transformer models in the last layer, the E-Z Reader and the BNC baseline for different word predictability scores computed with a 5-gram Kneser-Ney language model \cite{kneser1995improved, 41880}. Figure \ref{fig:cloze} shows the results on SST and Wikipedia for equally sized bins of word predictability scores. We can see that the Transformer models correlate better for more predictable words on both datasets whereas the E-Z Reader is less influenced by word predictability and already shows medium correlation on the most hard-to-predict words ($0.3-0.4$ for both, SST and Wikipedia). In fact, on SST, Transformers only pass the E-Z Reader on the most predictable tokens (word predictability $>0.03$).

We also compare correlations to human fixations based on the top-6 (most tokens) Part-of-speech (POS) tags. On SST, correlations with E-Z~Reader are very consistent across POS tags whereas attention flow shows weak correlations on proper nouns ($0.12$), nouns ($0.16$) and verbs ($0.16$) as presented in Figure \ref{fig:pos}. The BNC frequency baseline correlates well with human fixations on adpositions (ADP) which both assign comparably low values. Proper nouns (PROPN) are  overestimated in BNC as a result of their infrequent occurrence.

\begin{figure}[h!] 
\vspace{-3mm}
\includegraphics[width=1.0\linewidth]{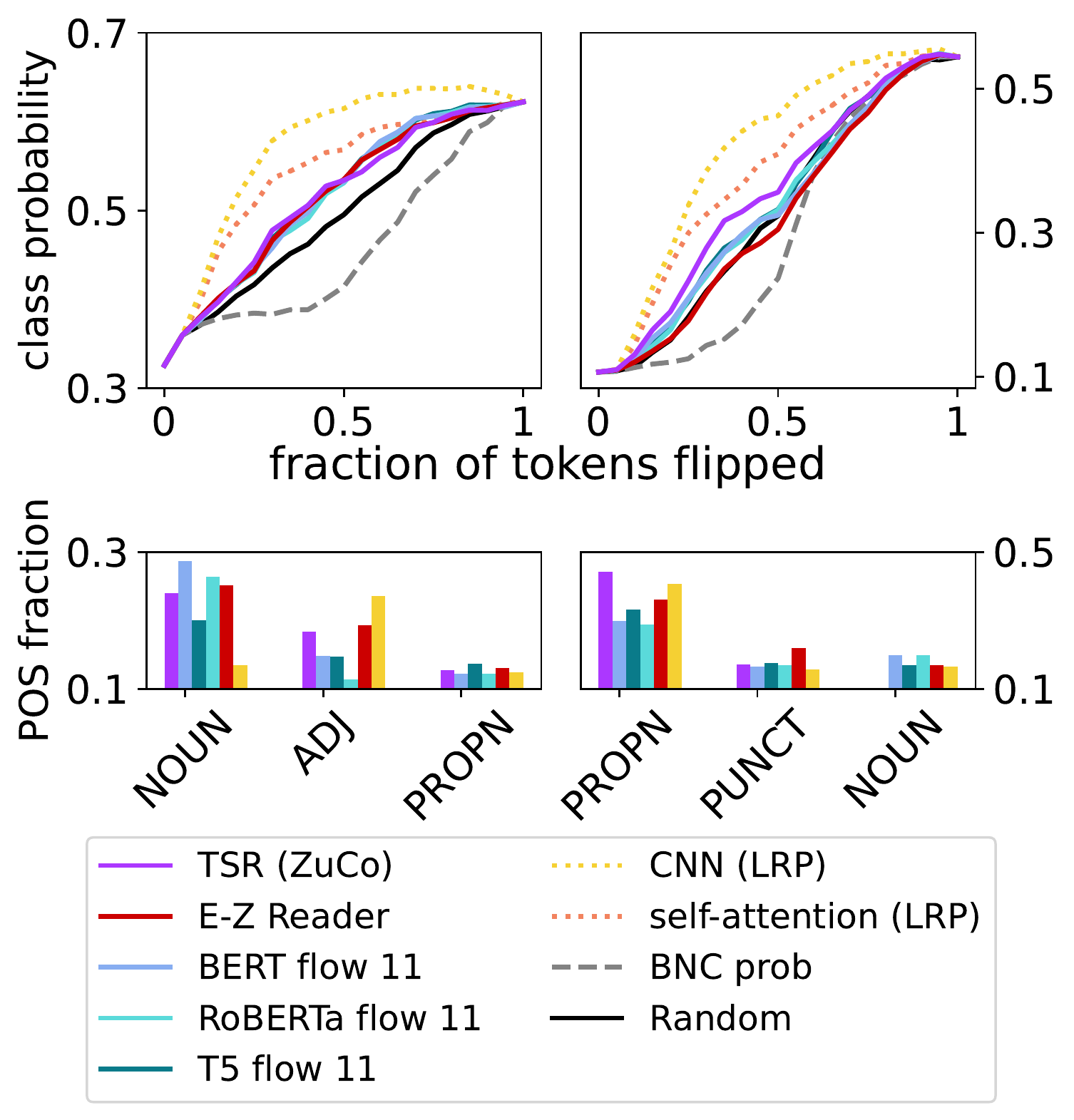}
\caption{Results of our reduction analysis where most important tokens are selected and fed into fine-tuned BERT models for sentiment classification \textit{(left)} and relation extraction \textit{(right)}. \textit{Upper:} we gradually measure output probability for the true label. Higher area under the curve reflects a stronger model sensitivity to adding important tokens. \textit{Lower:} Fractions of Most-selected POS tags at the first flip are displayed for human attention (TSR), flow 11, E-Z and BNC token probability.}
\label{fig:flipping}
\vspace{-3mm}
\end{figure}

\paragraph{Input reduction} 
When comparing machines to humans we typically regard the psychophysical data as the gold standard. We will now take the model perspective and test fidelity of both human and model attention patterns in task-tuned models. By this we aim to test how effective the exact token ranking based on attention scores is at producing the correct output probability.
We perform such an input reduction analysis \cite{feng-etal-2018-pathologies} using fine-tuned BERT models for both sentiment classification and relation extraction as the reference model and present results in Figure \ref{fig:flipping}. In our analysis, we observe - as to be expected - that adding tokens according to token probability (BNC prob)
performs even worse than randomly adding tokens. From-scratch trained models (CNN and self-attention) are most effective in selecting task-relevant tokens, and even more so than using any Transformer attention flow. Adding tokens based on human attention is as effective for the sentiment task as the E-Z Reader. Interestingly, for the relation extraction task, human attention vectors provide the most effective flipping order after the relevance-based shallow methods. All Transformer-based flows perform comparably in both tasks. To better understand what drives these effects we extract the fraction of POS tags for the first added token (see Figure \ref{fig:flipping} and full results in the Appendix Figure \ref{app:pos_reduction_analysis_full}). For sentiment reading, the flipping according to CNN relevances puts more emphasis on adjectives (ADJ) whereas the other methods tend to flip nouns (NOUN) first. Across the Transformer models RoBERTa relies much less on adjectives than any other model. In the relation extraction task, we observe that proper nouns (PROPN) are dominant (and adjectives play almost no role) in all model systems which highlights the role of task nature on the importance assignment. In addition, we see that the E-Z Reader overestimates the importance of punctuation, whereas proper nouns are least dominant in comparison to the other models. 

\begin{table}[t!]
\centering
\scriptsize
\setlength{\tabcolsep}{0.24em}
\begin{center}
\begin{tabular}{lccccccccccc}
\toprule
&
    \rotatebox{90}{TSR (ZuCo)} &     \rotatebox{90}{E-Z Reader} &   \rotatebox{90}{BNC inv prob} &      \rotatebox{90}{CNN (LRP)} & \rotatebox{90}{\pbox{1.8cm}{self-attention (LRP)}} &   \rotatebox{90}{BERT flow 11} & \rotatebox{90}{RoBERTa flow 11} &     \rotatebox{90}{T5 flow 11} &      \rotatebox{90}{BERT mean} &   \rotatebox{90}{RoBERTa mean} &        \rotatebox{90}{T5 mean} \\ 
\midrule SR & \cellcolor[HTML]{81AFCB} 3.44 & \cellcolor[HTML]{81AFCB} 3.44 & \cellcolor[HTML]{81B0CD} 3.40 & \cellcolor[HTML]{8BBDDA} 2.93 & \cellcolor[HTML]{BBD4E7} 2.16 & \cellcolor[HTML]{81ACC6} 3.57 & \cellcolor[HTML]{81ACC5} 3.61 & \cellcolor[HTML]{81ABC4} 3.61 & \cellcolor[HTML]{ABCEE3} 2.37 & \cellcolor[HTML]{9AC7DF} 2.65 & \cellcolor[HTML]{A6CCE2} 2.45 \\ 
TSR & \cellcolor[HTML]{81B1CD} 3.38 & \cellcolor[HTML]{81AFCA} 3.46 & \cellcolor[HTML]{81B1CD} 3.39 & \cellcolor[HTML]{89BCD9} 2.98 & \cellcolor[HTML]{FFFBFD} 1.81 & \cellcolor[HTML]{81ADC7} 3.54 & \cellcolor[HTML]{81ACC5} 3.60 & \cellcolor[HTML]{81ABC4} 3.63 & \cellcolor[HTML]{A4CBE1} 2.48 & \cellcolor[HTML]{9FC9E0} 2.56 & \cellcolor[HTML]{B0D0E4} 2.29 \\
\bottomrule
\end{tabular}
\caption{Mean entropy over all sentences for each task setting. Lower entropy means sparser token importance. The maximal entropy of a uniform model is 4.09 bits.}
\label{tab:entropy}
\end{center}
\vspace{-15px}
\end{table}

\paragraph{Entropy levels of Transformer flow is similar to those in human attention}
Averaged sentence-level entropy values on both datasets reveal that BERT, RoBERTa and T5 attention flow, the E-Z Reader and BNC obtain similar levels of sparsity as human attention around 3.4-3.6 bits  as summarized in Table \ref{tab:entropy}. 
Entropies are lower for the shallow networks with  self-attention (LRP) at 1.8-2.2 bits and CNN (LRP) at around 2.9 bits.
This difference in sparsity levels might explain the advantage of CNN and shallow self-attention in the input reduction analysis: Early addition of few but very relevant words has a strong effect on the model's decision compared to less sparse scoring as, e.g. in Transformers.
The shallow models were also trained from-scratch for the respective tasks whereas all other models (including human attention) are heavily influenced by a more general modeling of language which could explain attention to be distributed more broadly over all tokens.

\begin{table}[h!]
\centering
\small
\vspace{-1mm}
\setlength{\tabcolsep}{0.32em}

\begin{tabular}{p{0.6cm}ccccccccccc}
\toprule
 & \rotatebox{90}{BERT mean} & \rotatebox{90}{RoBERTa mean} & \rotatebox{90}{T5 mean} & \rotatebox{90}{fine-BERT mean } & \rotatebox{90}{T5 flow 11} & \rotatebox{90}{RoBERTa flow 11} & \rotatebox{90}{BNC inv prob} & \rotatebox{90}{E-Z Reader} & \rotatebox{90}{fine-BERT flow 11} & \rotatebox{90}{BERT flow 11} & \rotatebox{90}{ZuCo NR} \\
\midrule
NR  & .12 & .09 & .16 & .15 & .48 & .52 & .58 & .57 & .67 & .69 & - \\
TSR & .12 & .14 & .20 & .23 & .45 & .48 & .49 & .53 & .61 & .62 & .72 \\
\bottomrule
\end{tabular}
\caption{Correlations between human fixations and models on 48 duplicates appearing in the ZuCo dataset for both natural reading (NR) and relation extraction (task-specific reading - TSR). 
}
\label{tab:natural_reading}
\vspace{-3mm}
\end{table}

\paragraph{Natural reading versus task-specific reading} A unique feature of the ZuCo dataset is that it contains a subset of sentences that were presented to participants both in a task-specific (relation extraction) and a natural reading setting. This allows for a direct comparison of how correlation strength is influenced by the task. In Table \ref{tab:natural_reading} correlations of human gaze-based attention with model attentions are shown. The highest correlation can be observed when comparing human attention for task-specific and natural reading ($0.72$). The remaining model correlations correspond to the ranking and correlation strength observed in the main result (see Figure \ref{fig:spearman}). We observe lower correlation scores for the task-specific reading as compared to normal reading among attention flow, the E-Z~Reader and BNC. This suggests that these models capture the statistics of natural reading - as is expected for a cognitive model designed to the natural reading paradigm - and that task-related changes in human fixation patterns  are not reflected in Transformer attention flows. Interestingly, averaged last layer attention heads show a reverse effect (but at much weaker correlation strength). This might suggest that pre-training in Transformer models induces specificity of later layer attention heads to task-solving instead of general natural reading patterns.

\section{Related Work}\label{sec:relatedwork}

\paragraph{Saliency modeling}
Early computational models of visual attention have used bottom-up approaches to model the neural circuitry representing pre-attentive selection processes from visual input \cite{KochUllman85} and later the central idea of a saliency map was introduced \cite{Niebur_Koch96b}. 
A central hypothesis studying eye movements under task conditions is known as Yarbus theorem stating that a task can be directly decoded from fixation patterns \cite{Yarbus1967} which has found varying support  \cite{greene2012, Henderson2013, PMID:24665092}.

More recently, extracting features from deep pre-trained filters in combination with readout networks has boosted performance on the saliency task  \cite{kummerer_deepgaze_2016}. This progress has enabled modeling of more complex gaze patterns, e.g. vision-language tasks such as image captioning \cite{DBLP:journals/corr/SuganoB16}, visual question answering \cite{das-etal-2016-human} or text-guided object detection \cite{DBLP:conf/cvpr/VasudevanDG18}.

Predicting text gaze patterns has been studied extensively, often in the context of probabilistic  \cite{FENG200670, hara-etal-2012-predicting, matthies-sogaard-2013-blinkers, hahn-keller-2016-modeling}  or token transition models \cite{NilssonN09-0, HAJIABOLHASSANI2014127, courot2017}. More recently deep language features have been used as feature extractors in modeling text saliency \cite{sood-etal-2020-interpreting, hollenstein2021multilingual} opening the question of their cognitive plausibility.

\paragraph{Eye-tracking signals for NLP}
Augmenting machine learning models using human gaze information has been shown to improve performance for a number of different settings: Human attention patterns as regularization during model training have resulted in comparable or improved task performance in tagging part-of-speech \cite{barrett-sogaard-2015-reading, barrett-sogaard-2015-using, barrett-etal-2018-sequence}, sentence compression \cite{klerke-etal-2016-improving}, detecting sentiment \cite{mishra-etal-2016-leveraging,mishra-etal-2017-learning} or reading comprehension \cite{malmaud-etal-2020-bridging}. In these works, general free-viewing gaze data is used without consideration of the specific training task which opens the question of task-modulation in human reading.

\paragraph{From natural to task-specific reading} 
Recent work on reading often analyses eye-tracking data in combination with neuroimaging techniques such as EEG \cite{wenzel2017real} and f-MRI \cite{hillen2013identifying, choi2014neural}. Research questions thereby focus either on detecting relevant parts in text \cite{loboda2011inferring, wenzel2017real} or the difference between natural and pseudo-reading, i.e., text without syntax/semantics \cite{hillen2013identifying} or pseudo-words \cite{choi2014neural}. To the best of our knowledge there has not been any work on comparing fixations between natural reading and task-specific reading on classical NLP tasks such as relation extraction or sentiment classification.

\section{Discussion and Conclusion}\label{sec:disc}
In this paper, we have compared attention and relevance mechanisms of a wide range of models to human gaze patterns when solving sentiment classification on SST movie reviews and relation extraction on Wikipedia articles. 
We generally found that Transformer architectures are competitive with the {E-Z~Reader}, but only when computing attention flow scores.
We generally saw weaker correlations for relation extraction on Wikpedia, presumably due to simpler sentence structures and the occurrence of polarity words.
In the following, we discuss implications of our findings on NLP and Cognitive Science in more detail.

\paragraph{Lessons for NLP}
One implication of the above for NLP follows from the importance of attention flow in our experiments: Using human gaze to regularize or supervise attention weights has proven effective in previous work (\S\ref{sec:relatedwork}), but we observed that correlations with task-specific human attention increase significantly by using layer-dependent attention flow compared to using raw attention weights. This insight motivates going beyond regularizing raw attention weights or directly injecting human attention vectors during training, to instead optimize for correlation between attention flow and human attention. Jointly modeling language and human gaze has recently shown to yield competitive performance on paraphrase generation and sentence compression while resulting in more task-specific attention heads \cite{NEURIPS2020_460191c7}. For this study natural gaze patterns were also simulated using the E-Z~Reader.

Another potential implication concerns interpretability. It remains an open problem how best to interpret self-attention modules \cite{jain-wallace-2019-attention,wiegreffe-pinter-2019-attention}, and whether they provide meaningful explanations for model predictions.  
Including gradient information to explain Transformers has recently been considered to improve their interpretability \cite{Chefer_2021_CVPR, Chefer_2021_ICCV, DBLP:journals/corr/abs-2202-07304}.
A successful explanation of a  machine learning model should be faithful,  human-interpretable and practical to apply \cite{XAIreview2021}. Faithfulness and practicality is often evaluated using automated procedures such as input reduction experiments or measuring time and model complexity. By contrast, judging human-interpretability  typically requires costly experiments in well-controlled settings and obtaining human gold-standards for interpretability remain difficult \cite{MILLER20191, schmidt2019}. Using gaze data to evaluate the faithfulness and trustworthiness of machine learning models is a promising approach to increase model transparency. 

\paragraph{Lessons for  Cognitive Science} 
Attention flow in Transformers, especially for BERT models, correlates surprisingly well with human task-specific reading, but what does this tell us about the shortcomings of our cognitive models? We know that word frequency and semantic relationships between words influence word fixation times \cite{rayner1998eye}.
In our experiments, we see relatively high correlation between human fixations and the inverse word probability baseline which raises the question to what extent reading gaze is driven by low-level patterns such as word frequency or syntactic structure in contrast to more high-level semantic context or wrap-up effects.

In computer vision, cognitively inspired bottom-up models, e.g., using intensity and contrast features, are able to explain at most half of the gaze fixation information in comparison to the human gold standard \cite{kummerer2017low}. The robustness of the E-Z Reader on movie reviews is likely due to its explicit modeling of low-level properties such as word frequency or sentence length. 
BERT was recently shown to be primarily modeling higher-order word co-occurrence statistics \cite{sinha2021masked}. We argue that while Transformers are limited, e.g., in not capturing the dependency of human gaze on word length \cite{doi:10.1080/09541440340000213}, cognitive models seem to underestimate the role of word co-occurrence statistics. 

During reading, humans are faced with a trade-off between the precision of reading comprehension and reading speed, by avoiding unnecessary fixations \cite{hahn-keller-2016-modeling}. This trade-off is related to the input reduction experiments performed in Section \ref{sec:analyses}. Here, we observe that shallow methods score well at being sparse and effective in changing model output towards the correct class, but produce only weak correlation to human reading patterns when compared to layered language models. In comparison, extracted attention flow from pre-trained Transformer models correlates much better with human attention, but offers less sparse token attention. In other words, our results show that task-specific reading is sub-optimal relative to solving tasks and heavily regularized by natural reading patterns (see also our comparison of task-specific and natural reading in Section \ref{sec:analyses}).

\paragraph{Conclusion} In our experiments, we first and foremost found that Transformers, and especially BERT models, are competitive to the E-Z~Reader in terms of explaining human attention in task-specific reading. For this to be the case, computing attention flow scores (rather than raw attention weights) is important. Even so, the E-Z~Reader remains better at hard-to-predict words and is less sensitive to part of speech. While Transformers thus have some limitations compared to the {E-Z~Reader}, our results indicate that cognitive models have placed too little weight on high-level word co-occurrence statistics. 
Generally, Transformers and the {E-Z}~Reader correlate much better with human attention than other, shallow from-scratch trained sequence labeling architectures. Our input reduction experiments suggest that in a sense, both pre-trained language models {\em and} humans have suboptimal, i.e., less sparse,  task-solving strategies, and are heavily regularized by what is optimal in natural reading contexts.

\section*{Acknowledgements}
This  work  was partially funded  by  the  German  Ministry  for  Education  and  Research  as  BIFOLD  –  Berlin  Institute  for  the Foundations  of  Learning  and  Data  (ref.  01IS18025A  and ref.  01IS18037A), as well as by the Platform Intelligence in News project, which is supported by Innovation Fund Denmark via the Grand Solutions program. We thank Mostafa Abdou for fruitful discussions and Heather Lent, Miryam de Lhoneux and Vinit Ravishankar for proof-reading and valuable inputs on the manuscript.

\bibliography{main}
\bibliographystyle{acl_natbib}

\appendix
\label{sec:appendix}

\section{Model and Optimization Details}\label{app:model_optimization}
In the following we present details for all modes and describe the training details used for task-tunning. Model performance over five runs is reported in Table \ref{tab:accuracies}.
\begin{table*}
\begin{center}
\begin{tabular}{lll|ll}
\toprule
 &    Acc (SR) &     F1 (SR) &   Acc (REL) &    F1 (REL) \\
\midrule
self-attention       &  $69.0\pm0.2$ &  $64.5\pm2.2$ &  $67.5\pm1.3$ &  $55.5\pm2.0$ \\
CNN            &  $71.3\pm0.2$ &  $69.8\pm1.7$ &  $74.0\pm1.9$ &  $68.7\pm4.8$ \\
BERT-base      &  $76.0\pm0.1$ &  $67.0\pm3.0$ &  $78.3\pm1.5$ &  $72.7\pm3.3$ \\
BERT-large     &  $76.4\pm0.1$ &  $63.8\pm1.3$ &  $78.9\pm2.3$ &  $71.0\pm2.7$ \\
\bottomrule
\end{tabular}
\caption{Accuracy and F1 scores after fine-tuning on the respective task dataset over five runs: sentiment reading on SST (SR) and relation extraction on Wikipedia (REL). Samples that overlap with the ZuCo dataset were filtered out.}
\label{tab:accuracies}
\end{center}
\end{table*}
\subsection{CNN}
The CNN models use 300-dimensional pre-trained GloVe\_840B \cite{pennington2014glove} embeddings. Input sentences are tokenized using the SpaCy tokenizer \cite{spacy}. We use 150 convolutional filters of filter sizes $s=[3,4,5]$ with ReLu activation, followed by a max-pooling-layer and apply dropout of $p=0.5$  of the linear classification layer during training. 
For training we use a batchsize of $bs=50$ and train all model parameters using the Adam optimizer with a learning rate of $lr=1e-4$ for a maximum number of $T=20$ epochs. For all model trainings, we apply early stopping to avoid overfitting  during training and stop optimization as soon as the validation loss begins to increase.
To compute LRP relevances we use the general formulation of LRP propagation rules with  $\gamma=0.$ for the linear readout layers \cite{Montavon2019}. We take absolute values over resulting relevance scores since we find they correlate best with human attention in comparison to raw and rectified processing. For propagation through the max-pooling layer we apply the winner-take-all principle and for convolutional layers we use the LRP-$\gamma$ redistribution rule  and select $\gamma=0.5$ after a search over $\gamma=[0.,0.25,0.5,0.75,1.0]$ resulting in largest correlations to human attention.

\subsection{Self-Attention model}
For the multi-head self-attention model again use  300-dimensional pre-trained GloVe\_840B embeddings and tokenized via SpaCy. The architecture consists of a set of $k=3$ self-attention heads for the SR task and $k=8$ for REL. The resulting sentence representation is then fed into a linear classification readout layer with  $\gamma=0.$ and  which we also use for the propagation to input embeddings. During optimization we use $lr=1e-4$, $bs=50$ and $T=50$.

\subsection{Transformer Models}
We use standard BERT-base/large-uncased architectures and tokenizers as provided by the huggingface library \cite{wolf-etal-2020-transformers}. For BERT-base fine-tuning we use $lr=1e-5$ for REL and  $lr=1e-6$ for SR, $bs=32$ and $T=50$ for both tasks. For BERT-large we use $lr=1e-5$ for REL and $lr=5e-7$ for SR, $bs=16$ and $T=50$. 
For RoBERTa and T5 we use the RoBERTa-base and T5-base checkpoints and respective tokenizers.

\subsection{E-Z Reader}
We use version 10.2 of the E-Z~Reader with default parameters and 1000 repetitions.
Cloze scores, i.e.~word predictability scores, were therefore computed  using a 5-gram \textbf{Kneser-Ney} language model \cite{kneser1995improved} as provided by the  SRI Language Modeling Toolkit \cite{Stolcke02srilm} and  trained on the 1 billion token dataset \cite{41880}. Resulting perplexity on the held-out test set was $ppl=81.9$. Then, word-based total fixation times are computed from the E-Z~Readers trace files and averaged over all subjects. 

\section{Spearman versus Pearson correlation on sentence and token level}\label{app:spearman_pearson}
In addition to Spearman correlation over all tokens, we also report Pearson correlation coefficients on a sentence and token-level. Results are displayed in Table \ref{tab:apdx_corr_heatmap}. Compared to Spearman correlation on all tokens, the ranking does hardly change for Pearson or sentence-level correlations. Absolute correlation coefficients are higher for Spearman compared to Pearson and also are slightly higher on the sentence-level as compared to the token-level analysis. Biggest changes occur in a drop for BNC when Spearman correlation is calculated on all tokens for relation extraction and an increase for self-attention (LRP) in sentiment reading. We hypothesize that both effects can be traced back to the level of sparsity and the corresponding ranking for Spearman correlations. In our entropy analysis we found that, i.e. self-attention shows a sparser representation which was likely caused by the over-confidence of the model, and which could explain the  higher rank-based correlation.

\begin{table*}
    \centering
    \resizebox{\textwidth}{!}{
    \input{tables/acl_corrs}
    }
    
\caption{Full correlation  analysis  for   sentiment  reading \textit{(left)} and   relation  extraction \textit{(right)}.  We  show  Spearman and Pearson correlation coefficients between human fixations and models. Correlation coefficients were calculated per sentence and averaged (sen) or after concatenation of all sentences (tok)} 
\label{tab:apdx_corr_heatmap}
\end{table*}

\section{Input reduction - POS tag analysis}\label{app:pos_reduction}
\begin{figure}[!ht]
    \centering
    \includegraphics[width=.5\textwidth]{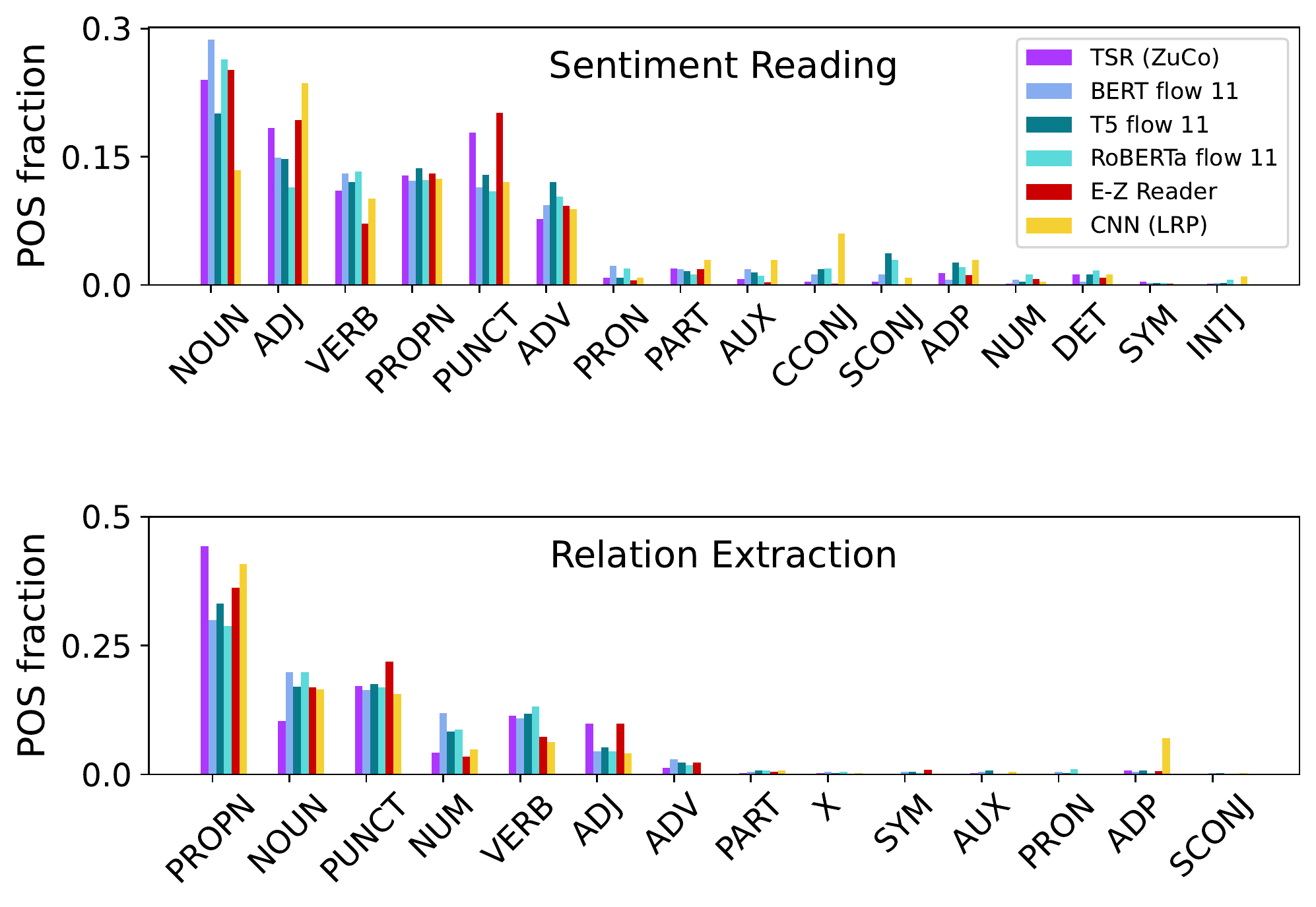}
    \caption{Full distribution of POS tags of most important first flip tokens for the task of sentiment reading (top) and relation extraction (bottom).}
    \label{app:pos_reduction_analysis_full}
\end{figure}
Figure \ref{app:pos_reduction_analysis_full} shows the full distribution of POS tags of the first tokens flipped. This extends Figure \ref{fig:flipping} where we only show the first 3 POS tags.

\section{Entropy analysis}\label{app:entropy}

We compute entropy values for different attention and relevance scores in both  task settings. To compensate for different sentence lengths we perform a stratified analysis such that every sentence length occurs equally often in both tasks. Sentence lengths which merely occur in one of the two tasks, are excluded from the sampling. Maximum entropy is reached for uniformly distributed token scores.

\end{document}

%% file: tables/acl_corrs.tex
\begin{tabular}{lrrrr|rrrrr}
\toprule
                         & \multicolumn{4}{c}{SR} & \multicolumn{4}{c}{TSR} \\
                         & \multicolumn{2}{c}{tok} & \multicolumn{2}{c}{sent} & \multicolumn{2}{c}{tok} & \multicolumn{2}{c}{sent} \\
                         &                         pearson &                        spearman &                         pearson &                        spearman &                         pearson &                        spearman &                         pearson &                        spearman \\
\midrule
            BNC inv prob &  \cellcolor[HTML]{B4DE81}  0.57 &  \cellcolor[HTML]{A0D169}  0.66 &  \cellcolor[HTML]{A9D774}  0.62 &  \cellcolor[HTML]{A5D46F}  0.64 &  \cellcolor[HTML]{DBF0BE}  0.34 &  \cellcolor[HTML]{CEEAAA}  0.41 &  \cellcolor[HTML]{C9E8A1}  0.45 &  \cellcolor[HTML]{C5E69B}  0.46 \\
               CNN (LRP) &  \cellcolor[HTML]{ECF5DD}  0.17 &  \cellcolor[HTML]{E6F5D0}  0.27 &  \cellcolor[HTML]{E6F5D1}  0.27 &  \cellcolor[HTML]{E7F5D3}  0.26 &  \cellcolor[HTML]{EEF6E3}  0.13 &  \cellcolor[HTML]{EEF5E2}  0.14 &  \cellcolor[HTML]{EAF5D9}  0.21 &  \cellcolor[HTML]{EBF5DC}  0.18 \\
          self-attention &  \cellcolor[HTML]{D7EEB8}  0.36 &  \cellcolor[HTML]{C3E698}  0.48 &  \cellcolor[HTML]{CCEAA7}  0.43 &  \cellcolor[HTML]{BAE28A}  0.54 &  \cellcolor[HTML]{E6F5D0}  0.27 &  \cellcolor[HTML]{C1E595}  0.49 &  \cellcolor[HTML]{C9E8A1}  0.44 &  \cellcolor[HTML]{ABD977}  0.61 \\
    self-attention (LRP) &  \cellcolor[HTML]{F2F6ED}  0.07 &  \cellcolor[HTML]{D2ECB0}  0.39 &  \cellcolor[HTML]{DBF0BE}  0.34 &  \cellcolor[HTML]{CCEAA7}  0.43 &  \cellcolor[HTML]{F1F6EA}  0.09 &  \cellcolor[HTML]{DEF1C4}  0.31 &  \cellcolor[HTML]{E6F5D0}  0.28 &  \cellcolor[HTML]{D7EEB8}  0.36 \\
             BERT flow 0 &  \cellcolor[HTML]{BCE28D}  0.52 &  \cellcolor[HTML]{A7D671}  0.62 &  \cellcolor[HTML]{ABD977}  0.61 &  \cellcolor[HTML]{A5D46F}  0.63 &  \cellcolor[HTML]{C5E69B}  0.47 &  \cellcolor[HTML]{B6E084}  0.55 &  \cellcolor[HTML]{B6E084}  0.55 &  \cellcolor[HTML]{ADDA79}  0.60 \\
             BERT flow 5 &  \cellcolor[HTML]{BAE28A}  0.53 &  \cellcolor[HTML]{A9D774}  0.61 &  \cellcolor[HTML]{ABD977}  0.60 &  \cellcolor[HTML]{A9D774}  0.61 &  \cellcolor[HTML]{C1E595}  0.49 &  \cellcolor[HTML]{BAE28A}  0.53 &  \cellcolor[HTML]{B2DD7F}  0.57 &  \cellcolor[HTML]{B2DD7F}  0.57 \\
            BERT flow 11 &  \cellcolor[HTML]{B8E187}  0.54 &  \cellcolor[HTML]{A9D774}  0.62 &  \cellcolor[HTML]{A9D774}  0.62 &  \cellcolor[HTML]{A7D671}  0.63 &  \cellcolor[HTML]{BEE390}  0.51 &  \cellcolor[HTML]{B4DE81}  0.56 &  \cellcolor[HTML]{ABD977}  0.60 &  \cellcolor[HTML]{ABD977}  0.61 \\
        fine-BERT flow 0 &  \cellcolor[HTML]{BCE28D}  0.52 &  \cellcolor[HTML]{A7D671}  0.62 &  \cellcolor[HTML]{ABD977}  0.61 &  \cellcolor[HTML]{A5D46F}  0.63 &  \cellcolor[HTML]{C5E69B}  0.47 &  \cellcolor[HTML]{B6E084}  0.55 &  \cellcolor[HTML]{B6E084}  0.55 &  \cellcolor[HTML]{ADDA79}  0.60 \\
        fine-BERT flow 5 &  \cellcolor[HTML]{BAE28A}  0.53 &  \cellcolor[HTML]{ABD977}  0.61 &  \cellcolor[HTML]{ADDA79}  0.59 &  \cellcolor[HTML]{ABD977}  0.61 &  \cellcolor[HTML]{C0E493}  0.50 &  \cellcolor[HTML]{B8E187}  0.54 &  \cellcolor[HTML]{B0DB7C}  0.59 &  \cellcolor[HTML]{ADDA79}  0.59 \\
       fine-BERT flow 11 &  \cellcolor[HTML]{B8E187}  0.54 &  \cellcolor[HTML]{A9D774}  0.62 &  \cellcolor[HTML]{A9D774}  0.62 &  \cellcolor[HTML]{A7D671}  0.63 &  \cellcolor[HTML]{BEE390}  0.51 &  \cellcolor[HTML]{B4DE81}  0.56 &  \cellcolor[HTML]{ABD977}  0.60 &  \cellcolor[HTML]{ABD977}  0.60 \\
       BERT-large flow 0 &  \cellcolor[HTML]{BEE390}  0.51 &  \cellcolor[HTML]{A9D774}  0.61 &  \cellcolor[HTML]{A9D774}  0.62 &  \cellcolor[HTML]{A7D671}  0.63 &  \cellcolor[HTML]{C5E69B}  0.47 &  \cellcolor[HTML]{B8E187}  0.54 &  \cellcolor[HTML]{B2DD7F}  0.57 &  \cellcolor[HTML]{ADDA79}  0.60 \\
      BERT-large flow 11 &  \cellcolor[HTML]{B6E084}  0.55 &  \cellcolor[HTML]{A7D671}  0.63 &  \cellcolor[HTML]{A9D774}  0.62 &  \cellcolor[HTML]{A9D774}  0.62 &  \cellcolor[HTML]{C0E493}  0.50 &  \cellcolor[HTML]{B6E084}  0.55 &  \cellcolor[HTML]{B2DD7F}  0.57 &  \cellcolor[HTML]{B2DD7F}  0.57 \\
      BERT-large flow 23 &  \cellcolor[HTML]{B6E084}  0.55 &  \cellcolor[HTML]{A7D671}  0.63 &  \cellcolor[HTML]{A9D774}  0.62 &  \cellcolor[HTML]{A9D774}  0.62 &  \cellcolor[HTML]{C0E493}  0.50 &  \cellcolor[HTML]{B6E084}  0.55 &  \cellcolor[HTML]{B2DD7F}  0.57 &  \cellcolor[HTML]{B2DD7F}  0.57 \\
  fine-BERT-large flow 0 &  \cellcolor[HTML]{BEE390}  0.51 &  \cellcolor[HTML]{A9D774}  0.61 &  \cellcolor[HTML]{A9D774}  0.62 &  \cellcolor[HTML]{A7D671}  0.63 &  \cellcolor[HTML]{C5E69B}  0.47 &  \cellcolor[HTML]{B8E187}  0.54 &  \cellcolor[HTML]{B2DD7F}  0.57 &  \cellcolor[HTML]{ADDA79}  0.60 \\
 fine-BERT-large flow 11 &  \cellcolor[HTML]{B6E084}  0.55 &  \cellcolor[HTML]{A7D671}  0.63 &  \cellcolor[HTML]{A9D774}  0.62 &  \cellcolor[HTML]{A9D774}  0.62 &  \cellcolor[HTML]{C0E493}  0.50 &  \cellcolor[HTML]{B6E084}  0.55 &  \cellcolor[HTML]{B4DE81}  0.57 &  \cellcolor[HTML]{B2DD7F}  0.57 \\
 fine-BERT-large flow 23 &  \cellcolor[HTML]{B6E084}  0.55 &  \cellcolor[HTML]{A7D671}  0.63 &  \cellcolor[HTML]{A9D774}  0.62 &  \cellcolor[HTML]{A9D774}  0.62 &  \cellcolor[HTML]{C0E493}  0.50 &  \cellcolor[HTML]{B6E084}  0.55 &  \cellcolor[HTML]{B4DE81}  0.57 &  \cellcolor[HTML]{B2DD7F}  0.57 \\
          RoBERTa flow 0 &  \cellcolor[HTML]{CAE9A4}  0.44 &  \cellcolor[HTML]{B8E187}  0.54 &  \cellcolor[HTML]{BCE28D}  0.52 &  \cellcolor[HTML]{B8E187}  0.55 &  \cellcolor[HTML]{D9EFBB}  0.35 &  \cellcolor[HTML]{CAE9A4}  0.43 &  \cellcolor[HTML]{CCEAA7}  0.42 &  \cellcolor[HTML]{C5E69B}  0.47 \\
          RoBERTa flow 5 &  \cellcolor[HTML]{DEF1C4}  0.32 &  \cellcolor[HTML]{CCEAA7}  0.42 &  \cellcolor[HTML]{C9E8A1}  0.45 &  \cellcolor[HTML]{C7E79E}  0.46 &  \cellcolor[HTML]{E6F5D1}  0.26 &  \cellcolor[HTML]{DCF1C1}  0.33 &  \cellcolor[HTML]{D7EEB8}  0.36 &  \cellcolor[HTML]{D0EBAD}  0.40 \\
         RoBERTa flow 11 &  \cellcolor[HTML]{C9E8A1}  0.44 &  \cellcolor[HTML]{BEE390}  0.51 &  \cellcolor[HTML]{BEE390}  0.51 &  \cellcolor[HTML]{BCE28D}  0.52 &  \cellcolor[HTML]{D5EDB5}  0.37 &  \cellcolor[HTML]{D0EBAD}  0.41 &  \cellcolor[HTML]{C7E79E}  0.45 &  \cellcolor[HTML]{C7E79E}  0.46 \\
               T5 flow 0 &  \cellcolor[HTML]{C9E8A1}  0.44 &  \cellcolor[HTML]{BAE28A}  0.53 &  \cellcolor[HTML]{BEE390}  0.51 &  \cellcolor[HTML]{B8E187}  0.54 &  \cellcolor[HTML]{D5EDB5}  0.37 &  \cellcolor[HTML]{C9E8A1}  0.44 &  \cellcolor[HTML]{C3E698}  0.47 &  \cellcolor[HTML]{C0E493}  0.50 \\
               T5 flow 5 &  \cellcolor[HTML]{CCEAA7}  0.43 &  \cellcolor[HTML]{C0E493}  0.50 &  \cellcolor[HTML]{C1E595}  0.49 &  \cellcolor[HTML]{C0E493}  0.49 &  \cellcolor[HTML]{D9EFBB}  0.35 &  \cellcolor[HTML]{D2ECB0}  0.40 &  \cellcolor[HTML]{CAE9A4}  0.44 &  \cellcolor[HTML]{CAE9A4}  0.43 \\
              T5 flow 11 &  \cellcolor[HTML]{C9E8A1}  0.44 &  \cellcolor[HTML]{BEE390}  0.51 &  \cellcolor[HTML]{BEE390}  0.51 &  \cellcolor[HTML]{BAE28A}  0.53 &  \cellcolor[HTML]{D5EDB5}  0.37 &  \cellcolor[HTML]{CCEAA7}  0.42 &  \cellcolor[HTML]{C7E79E}  0.46 &  \cellcolor[HTML]{C5E69B}  0.46 \\
               BERT mean &  \cellcolor[HTML]{F4F6F0}  0.04 &  \cellcolor[HTML]{EEF5E2}  0.14 &  \cellcolor[HTML]{F1F6EA}  0.10 &  \cellcolor[HTML]{F0F6E6}  0.11 &  \cellcolor[HTML]{F7F4F6} -0.03 &  \cellcolor[HTML]{F0F6E6}  0.11 &  \cellcolor[HTML]{F5F6F3}  0.02 &  \cellcolor[HTML]{F1F6EA}  0.09 \\
         fine-BERT mean  &  \cellcolor[HTML]{F4F6F1}  0.03 &  \cellcolor[HTML]{F1F6EA}  0.09 &  \cellcolor[HTML]{F4F6F0}  0.05 &  \cellcolor[HTML]{F4F6F1}  0.03 &  \cellcolor[HTML]{F7F4F6} -0.03 &  \cellcolor[HTML]{F0F6E8}  0.10 &  \cellcolor[HTML]{F5F6F3}  0.02 &  \cellcolor[HTML]{F2F6EB}  0.08 \\
         BERT-large mean &  \cellcolor[HTML]{F7F5F6} -0.01 &  \cellcolor[HTML]{EAF5DA}  0.20 &  \cellcolor[HTML]{F0F6E8}  0.10 &  \cellcolor[HTML]{E6F5D0}  0.28 &  \cellcolor[HTML]{F7F4F6} -0.03 &  \cellcolor[HTML]{EEF5E2}  0.14 &  \cellcolor[HTML]{F7F6F6} -0.01 &  \cellcolor[HTML]{EEF5E2}  0.14 \\
   fine-BERT-large mean  &  \cellcolor[HTML]{F7F5F6} -0.02 &  \cellcolor[HTML]{F0F6E6}  0.11 &  \cellcolor[HTML]{F4F6F1}  0.04 &  \cellcolor[HTML]{ECF5DF}  0.17 &  \cellcolor[HTML]{F9EFF4} -0.09 &  \cellcolor[HTML]{F8F2F5} -0.05 &  \cellcolor[HTML]{F9ECF3} -0.12 &  \cellcolor[HTML]{F8F1F4} -0.06 \\
            RoBERTa mean &  \cellcolor[HTML]{E9F5D7}  0.22 &  \cellcolor[HTML]{E9F5D7}  0.22 &  \cellcolor[HTML]{E7F5D3}  0.26 &  \cellcolor[HTML]{EAF5D9}  0.21 &  \cellcolor[HTML]{F2F6EB}  0.08 &  \cellcolor[HTML]{F0F6E8}  0.10 &  \cellcolor[HTML]{EEF5E2}  0.14 &  \cellcolor[HTML]{F1F6EA}  0.10 \\
                 T5 mean &  \cellcolor[HTML]{F7F6F6} -0.00 &  \cellcolor[HTML]{F3F6EE}  0.06 &  \cellcolor[HTML]{F7F6F6} -0.00 &  \cellcolor[HTML]{F2F6ED}  0.07 &  \cellcolor[HTML]{F7F4F6} -0.02 &  \cellcolor[HTML]{F1F6EA}  0.10 &  \cellcolor[HTML]{F6F6F4}  0.02 &  \cellcolor[HTML]{EBF5DC}  0.19 \\
              E-Z Reader &  \cellcolor[HTML]{A5D46F}  0.64 &  \cellcolor[HTML]{A2D36C}  0.65 &  \cellcolor[HTML]{99CD61}  0.69 &  \cellcolor[HTML]{9ED066}  0.67 &  \cellcolor[HTML]{C7E79E}  0.46 &  \cellcolor[HTML]{BEE390}  0.51 &  \cellcolor[HTML]{B4DE81}  0.56 &  \cellcolor[HTML]{B4DE81}  0.56 \\
\bottomrule
\end{tabular}